\documentclass[conference]{IEEEtran}
\IEEEoverridecommandlockouts

\usepackage{cite}
\usepackage{amsmath,amssymb,amsfonts}
\usepackage{graphicx}
\usepackage{textcomp}
\usepackage{booktabs}
\usepackage{xcolor}
\usepackage{tikz}
\usetikzlibrary{arrows.meta,positioning,calc}
\usepackage[tracking=false]{microtype}
\usepackage[hidelinks]{hyperref}

\newif\ifanonymous
\anonymousfalse

\newif\ifappendix
\appendixtrue

\begin{document}

%
%
\newcommand{\releaseurl}{https://github.com/ParamThakkar123/sparse_world_models}

\title{Modeling What Changes: Sparse, Residual World Models\\ for Object-Centric Manipulation
\thanks{Code, data generators, checkpoints, and machine-readable result tables
\ifx\releaseurl\empty will be released upon publication\else are released at
\url{\releaseurl}\fi.}}

\ifanonymous
  \author{\IEEEauthorblockN{Anonymous Author(s)}
  \IEEEauthorblockA{Submission under double-blind review}}
\else
  \author{\IEEEauthorblockN{Param Thakkar}
  \IEEEauthorblockA{Veermata Jijabai Technological Institute\\\texttt{puthakkar\_b22@ce.vjti.ac.in}}
  \and
  \IEEEauthorblockN{Parsika Paresh Shah}
  \IEEEauthorblockA{Arizona State University\\\texttt{pshah144@asu.edu}}
  \and
  \IEEEauthorblockN{Manisha Sushant Gote}
  \IEEEauthorblockA{ZuiGO Private Limited\\\texttt{manisha@zuigo.ai}}}
\fi

\maketitle

\begin{abstract}
Monolithic world models predict the entire next state at every step, spending
capacity re-predicting the static majority of a scene and injecting error into it.
We ask whether explicitly modeling \emph{what changes} (a per-object change gate
plus a residual delta head that perturbs only gated objects, copying the rest
verbatim) is a more effective and interpretable bias. On a MuJoCo tabletop pushing
benchmark (3--8 objects), the sparse/residual model is 2.5--4.6$\times$ more
accurate than a dense MLP at 8.6--11.1$\times$ fewer parameters \emph{overall},
but this headline is driven entirely by error suppression on the static majority:
at $N{=}5$, overall $L_2$ is 0.104 (sparse) vs.\ 0.107 (no-op); at $N{=}8$, 0.081 vs.\
0.081; changed-object $L_2$ is 0.426 vs.\ 0.446 ($N{=}5$) and 0.466 vs.\ 0.470
($N{=}8$), i.e.\ at or barely above the floor of predicting no motion. The durable
win is therefore \emph{change detection that avoids error injection} (F1 0.80--0.87
vs.\ dense degenerate, precision 0.92--0.97) with delta regression at roughly no-op
quality, not accurate dynamics on movers. The gate transfers across counts (99.4\%
F1 retention) and is more sample-efficient. In open-loop rollout sparse compounds
far less error than dense but remains worse than no-op overall at every $N$ and
horizon; we report the decomposition and a mover-only rollout to make this explicit.
For planning, once featurized for planner-visited states, sparse is $0.23\pm0.06$
success vs.\ random $0.15$ (weakest seed exactly 0.15) and dense $0.00$ at every
seed: not significantly better than random, but separated from dense. Code, data
generators, and checkpoints are released at \url{https://github.com/ParamThakkar123/sparse_world_models}.
\end{abstract}


\section{Introduction}
Physical environments are overwhelmingly sparse in \emph{change}. When a robot
nudges one block on a cluttered table, a single object's pose changes and the rest
of the scene is exactly as it was. Yet the dominant recipe for learned world
models, encoding the scene and predicting the entire next state (or latent)
monolithically~\cite{ha2018world,hafner2020dream}, ignores this structure. Such a
model must reproduce every object at every step, which has two costs. First,
\emph{compute and capacity}: parameters and operations are spent re-predicting the
static majority rather than the few objects that matter. Second, and more
insidious, \emph{error injection}: a regressor that outputs all poses inevitably
perturbs objects that should have been copied verbatim, and in a closed-loop
rollout that error accumulates on precisely the objects that never moved, so the
predicted world drifts.

There is also an \emph{interpretability} gap: a monolithic predictor offers no explicit
handle on \emph{what} it thinks changed, entangling that belief in a single regression
output. For control and debugging, a model that names the objects it expects to move is
far more useful than one that silently re-renders everything.

\begin{figure}[t]
\centering
\resizebox{\columnwidth}{!}{%
\begin{tikzpicture}[
  font=\footnotesize,
  box/.style={draw, rounded corners, minimum height=6mm, inner sep=3pt, align=center},
  net/.style={box, fill=blue!8},
  op/.style={draw, circle, inner sep=1pt, minimum size=5mm},
  >={Stealth[length=2mm]}, node distance=5mm and 7mm]
  \node[box, fill=black!5] (in) {per-object\\features $\phi_i(s_t,a_t)$};
  \node[net, above right=3mm and 9mm of in] (gate) {gate net};
  \node[net, below right=3mm and 9mm of in] (delta) {delta head};
  \node[op, right=7mm of gate] (g) {$g_i$};
  \node[op, right=7mm of delta] (d) {$\Delta_i$};
  \node[op, right=11mm of $(g)!0.5!(d)$] (mult) {$\times$};
  \node[box, fill=green!10, right=6mm of mult] (out) {$\hat p_i^{t+1}=p_i^{t}+g_i\Delta_i$};
  \draw[->] (in) -- (gate);
  \draw[->] (in) -- (delta);
  \draw[->] (gate) -- (g);
  \draw[->] (delta) -- (d);
  \draw[->] (g) -- (mult);
  \draw[->] (d) -- (mult);
  \draw[->] (mult) -- (out);
  \node[above=1mm of g, font=\scriptsize\itshape] {change gate (STE)};
  \node[below=1mm of d, font=\scriptsize\itshape] {residual $\in\mathbb{R}^3$};
\end{tikzpicture}%
}
\caption{The sparse/residual head, applied per object with weights shared across
objects. A gate network emits a change decision $g_i$ (discrete, trained by a
straight-through estimator); a delta head emits a residual pose increment $\Delta_i$.
The next pose is the current pose plus the \emph{masked} residual, so objects the
gate leaves off are copied verbatim.}
\label{fig:arch}
\end{figure}
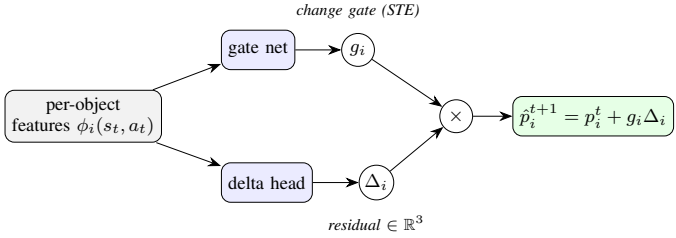

We ask whether explicitly modeling what changes addresses all three. Our model is
deliberately simple and object-centric: a per-object change \emph{gate} predicts
which objects move, and a residual \emph{delta head} predicts a pose increment
applied \emph{only} to gated objects; every other object is carried forward
unchanged (Fig.~\ref{fig:arch}). This sparse/residual factorization mirrors the
sparsity of physical interaction, shares parameters across objects, is independent
of object count, and exposes an explicit, inspectable change mask.

We evaluate on a procedurally generated MuJoCo tabletop pushing benchmark scaling
from 3 to 8 free objects, against a dense MLP state
predictor and a no-op (predict-no-change) baseline. Our contributions:
\begin{itemize}
  \item A sparse/residual object-centric world model (gate plus residual delta head)
  that is 2.5--4.6$\times$ more accurate than dense overall at 8.6--11.1$\times$
  fewer parameters, \emph{but} whose advantage is error suppression on the static
  majority: overall $L_2$ matches no-op at $N{=}5,8$ and changed-object $L_2$ is at
  or barely above no-op (Sec.~\ref{sec:prediction}, App.~\ref{app:oracle}); the
  publishable claim is a change detector that avoids injection, with delta regression
  at roughly no-op quality.
  \item Analyses isolating \emph{why}: a parameter-matched control, a
  capacity-matched ladder showing the \emph{gate} is the largest single step, an
  oracle-gate diagnostic (bottleneck is delta regression), and an explicit discussion
  of the F1 metric's definitional bias and a missing dense-residual rung
  (Sec.~\ref{sec:analysis}, App.~\ref{app:defs}).
  \item Evidence on world-model properties, stated honestly: rollout is far better
  than dense but uniformly \emph{worse} than no-op overall; decomposition into
  static vs.\ mover error and a mover-only rollout curve makes this explicit
  (Sec.~\ref{sec:worldmodel}). Transfer across counts (99.4\% retention) and sample
  efficiency hold.
  \item A planning study where sparse, after featurization for planner states, is
  $0.23\pm0.06$ vs.\ random $0.15$ and dense $0.00$ at every seed: not significantly
  better than random, but separated from dense at every seed
  (Sec.~\ref{sec:planning}).
\end{itemize}

\textbf{Scope.} A deliberately controlled study, whose bounds we state up front:
structured per-object state (poses and velocities), not pixels; 3 to 8 rigid boxes; small
MLPs ($<0.1$M parameters) that train, plan, and evaluate on a single laptop; and one
push-to-goal planning task. Section~\ref{sec:limitations} revisits the consequences.

\section{Related Work}\label{sec:related}
\textbf{Dense world models and model-based control.} Learned world models compress
observations into a latent state and predict its evolution for imagination, planning,
and policy learning~\cite{ha2018world,hafner2020dream}; sampling-based MPC
(PETS~\cite{chua2018pets}, MPPI~\cite{williams2017mppi},
CEM~\cite{rubinstein1999cem}) rolls action sequences through such a model. These predict
the full state or latent \emph{monolithically}, with no notion of which parts of the
scene are inert, so error spreads across the whole state.

\textbf{Object-centric and structured world models.} Graph- and object-factored dynamics
models share computation across entities and generalize across entity
counts~\cite{battaglia2016interaction,sanchez2020gns,kipf2020cswm}. Ours is in this
family (per-object features, shared weights, count-agnostic) but differs in \emph{what}
it predicts: an explicit per-object change \emph{decision} plus a residual, targeting the
sparsity of interaction, not only its relational structure.

\textbf{Sparse coding and conditional computation.} Good representations of natural
signals are \emph{sparse}~\cite{olshausen1996sparse}. Conditional-computation and sparsely-gated
architectures route each input through a small, input-dependent subset of the network via
learned discrete gates~\cite{shazeer2017moe,bengio2013conditional}, trained with
straight-through or relaxed-categorical
estimators~\cite{jang2017gumbel,maddison2017concrete}. We import this into the
\emph{output} of a dynamics model, making sparsity a property of the prediction itself
rather than of intermediate features.

\textbf{Pushing dynamics.} Pushing spans analytic mechanics~\cite{mason1986pushing},
real-world planar-pushing datasets~\cite{yu2016push}, and learned forward/inverse
models~\cite{agrawal2016poke,finn2017foresight}. Closest to ours,
SE3-Nets~\cite{byravan2017se3nets} predict per-object rigid-body motions with soft masks
from point clouds; we instead learn a \emph{discrete} change gate under an explicit
sparsity objective. \textbf{The gap.} Object-centric models supply relational
structure but still predict every object densely; conditional-computation methods supply
input-dependent sparsity but not over a world model's \emph{change structure}. We close
that gap and evaluate the full chain from prediction to control.


\section{Method}\label{sec:method}
\textbf{State and action.} The scene state $s_t$ concatenates the pusher position,
per-object planar pose $(x,y,\theta)$, per-object velocity, and the goal. Actions
$a_t$ are end-effector delta-$xy$ commands to a position-controlled pusher.

\textbf{Dense baseline.} A multilayer perceptron $f_\theta(s_t,a_t)$ regresses all
object poses at $t{+}1$ under an $L_2$ loss (hidden width 256, 3 layers).

\textbf{Sparse/residual model (Fig.~\ref{fig:arch}).} For each object $i$ we build a
feature vector $\phi_i(s_t,a_t)$ from its pose, velocity, relative goal and pusher
positions, the action, and a permutation-invariant aggregate over the other objects.
A gate network outputs a change logit; a delta head outputs a residual
$\Delta_i\in\mathbb{R}^3$. The predicted next pose is the current pose plus the
\emph{masked} residual,
\begin{equation}
  \hat{p}_i^{\,t+1} = p_i^{\,t} + g_i \,\Delta_i, \qquad g_i \sim \text{Gate}(\phi_i),
  \label{eq:pred}
\end{equation}
where $g_i\in\{0,1\}$ is a discrete gate sampled with a Gumbel straight-through
estimator~\cite{jang2017gumbel,bengio2013conditional} so the model trains end-to-end
while remaining hard (exactly zero update) at inference. Because the gate and delta
heads are shared across objects and no layer is sized to the object count, a model
trained at one count runs at any other.

\textbf{Loss.} We combine a class-balanced binary cross-entropy on the gate against
the ground-truth changed mask $m^\star$, an $L_2$ regression on the residual
supervised only on changed objects, and a sparsity penalty on the mean gate
activation $\bar g$:
\begin{equation}
  \mathcal{L} = \mathrm{BCE}(g, m^\star) + \lambda_\Delta\, m^\star \!\cdot\! \lVert \Delta - \Delta^\star \rVert_2^2 + \lambda_s\, \bar{g}.
  \label{eq:loss}
\end{equation}
The sparsity weight $\lambda_s$ trades gate recall for precision: a five-point sweep
shows precision rising 0.94 to 0.97 and recall falling 0.81 to 0.71 as $\lambda_s$
grows, with F1 degrading only gently. We use $\lambda_s{=}0.2$ for prediction.

\section{Experimental Setup}\label{sec:setup}
\textbf{Environment and data.} A procedural MuJoCo tabletop with $N$ free 5\,cm boxes
and a scripted pushing policy. For each $(N,\text{seed})$ we log $(s_t,a_t,s_{t+1})$
with ground-truth per-object changed mask and delta, filter to a hard subset (steps
with real motion) so metrics are not dominated by trivially static steps, and split
80/10/10 with an explicit configuration-leakage guard. \textbf{Baselines.} Dense MLP;
no-op (predict no change). \textbf{Metrics.} Overall and changed/unchanged per-object
pose $L_2$; change-detection precision, recall, and F1; parameters, operations,
latency. \textbf{Scaling.} $N\in\{3,5,8\}$; the 8-object layout uses wider bounds and
tighter spacing to fit the boxes (a stated caveat for cross-$N$ density trends).
Tables~\ref{tab:headline}--\ref{tab:planning} are mean plus or minus standard deviation
over three training seeds; single-seed (seed 0) analyses are flagged in place. \textbf{Hardware.} All models are small MLPs ($<0.1$M parameters) and run
on a single laptop (Intel Core i7-12650H CPU, NVIDIA RTX 3050 GPU, CUDA 12.6); either
device works via \texttt{-{}-device}. Latencies are CPU: at this scale per-call
GPU kernel-launch overhead dominates.
\textbf{Training.} Adam, learning rate $10^{-3}$, batch 128; 25 epochs dense, 15 sparse
(25 for the contact/planning variant). Gate and delta heads are each 2-layer width-128
MLPs; Gumbel temperature 1.0, delta term weighted by the predicted gate probability,
$\lambda_s{=}0.05$ for planning. Data: 250 scripted episodes $\times$ 100 steps per
$(N,\text{seed})$, hard-subset threshold 0.02\,m; the planning set adds 200 scripted
$\times$ 80 and 350 random $\times$ 60 episodes. CEM: elite fraction $0.1$, initial std
$0.6$, terminal weight $3.0$, proximity weight $0.3$, $\le$60 steps per episode.

\section{Prediction Accuracy}\label{sec:prediction}
\begin{figure}[t]
\centering
\includegraphics[width=\linewidth]{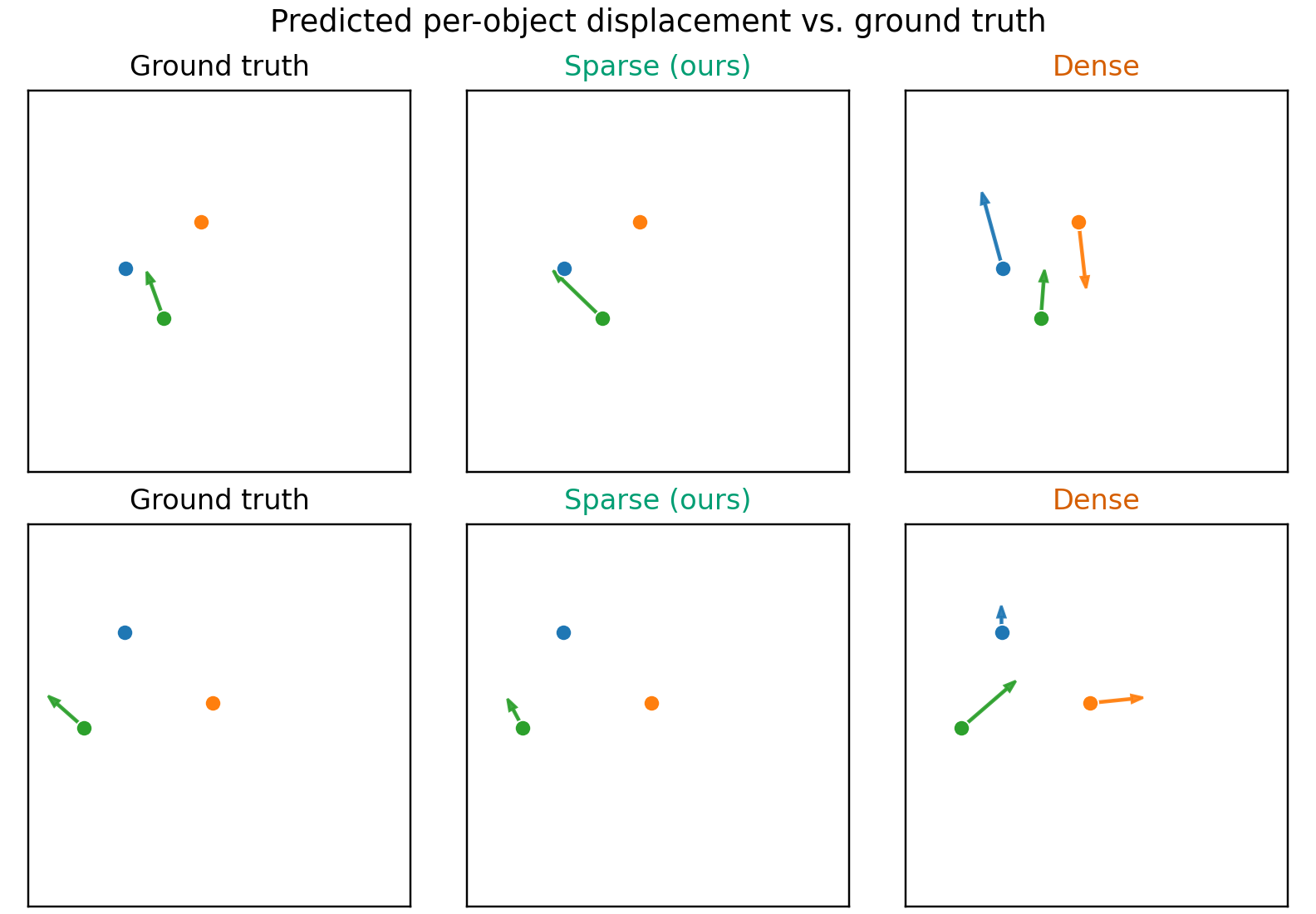}
\caption{Predicted per-object displacement (arrows from current to predicted next
position) on two example scenes. The sparse model (center) moves only the object that
truly moves, matching ground truth (left); the dense model (right) hallucinates motion
on objects that should stay put.}
\label{fig:qual}
\end{figure}

Table~\ref{tab:headline} reports the headline comparison. The sparse model's overall
$L_2$ is 2.5--4.6$\times$ lower than dense at every $N$ (8.6--11.1$\times$ fewer
parameters), but the framing ``predicts dynamics far more accurately'' overstates what
the numbers show. At $N{=}5$, sparse $L_2$ is $0.101\pm0.004$ vs.\ no-op $0.107$; at
$N{=}8$, $0.071\pm0.016$ vs.\ $0.081$ (Tab.~\ref{tab:ladder}); Table~I's 0.081 vs.\
0.081 is exact equality. On the objects that actually move, App.~\ref{app:oracle}
gives changed-object $L_2$ 0.466 (sparse) vs.\ 0.470 (no-op) at $N{=}8$ and 0.426 vs.\
0.446 at $N{=}5$: at or barely above the floor of predicting that nothing moves.
The mechanism in Sec.~\ref{sec:analysis} is correct and we state it plainly here:
the win is \emph{not} accurate delta regression, it is \emph{not corrupting the static
majority} ($L_2^{\text{unch}}$ $0.211\to0.001$ at $N{=}3$ via the gate), with delta
regression at roughly no-op quality. That is the publishable claim and we headline
it as such. Change detection sustains F1 0.80--0.87 (precision 0.92--0.97) where dense
is degenerate (recall $\approx1$); Fig.~\ref{fig:qual} shows sparse copying the rest
while dense perturbs the whole scene. A dense-interaction control restores $\sim12\times$
margin at $N{=}8$, but does not change the mover-level conclusion.

\begin{table}[t]
\caption{Prediction accuracy and efficiency (mean $\pm$ std, 3 seeds).}
\label{tab:headline}
\centering
\setlength{\tabcolsep}{3pt}
\begin{tabular}{@{}ccccc@{}}
\toprule
$N$ & sparse F1 & sparse $L_2$ & dense $L_2$ & params (dense/sparse) \\
\midrule
3 & $0.867 \pm 0.021$ & $0.136 \pm 0.046$ & $0.347 \pm 0.053$ & $11.1\times$ \\
5 & $0.802 \pm 0.024$ & $0.101 \pm 0.004$ & $0.319 \pm 0.007$ & $9.8\times$ \\
8 & $0.829 \pm 0.008$ & $0.071 \pm 0.016$ & $0.329 \pm 0.023$ & $8.6\times$ \\
\bottomrule
\end{tabular}
\end{table}


\textbf{Efficiency, honestly.} The win is \emph{parameters}, not wall-clock: per-object
heads scale with $N$, eroding the operation-count ratio ($3.8\times$ to $1.1\times$), and
dense's single matmul is faster in latency. We claim capacity efficiency, not speed.

\section{Why It Wins}\label{sec:analysis}
(All diagnostics in this section are seed 0, test split.)
\textbf{Not just smaller.} Shrinking the dense MLP to the sparse
parameter budget does not help it: at matched parameters sparse's overall $L_2$ is 3 to 5
times lower and its F1 far higher at every count, so the degeneracy is not a capacity
artifact.

\textbf{It is the gate, not just object-centricity.} The sparse model is both
object-centric \emph{and} change-modeling. A capacity-matched ladder (Tab.~\ref{tab:ladder})
adds one at a time: \textsc{oc-abs} (shared per-object MLP, absolute poses),
\textsc{oc-res} (always-applied residual), then gated; \textsc{oc} rungs match sparse
within $0.04\%$. Featurization earns 35--47\% of the $L_2$ gap, residual a little more,
but the gate is the largest single step (1.7--2.0$\times$) and the only one reaching
below no-op. All three ungated rungs share identical F1 (recall $=1$), but this is
partly definitional (App.~\ref{app:defs}): ungated masks are derived by thresholding
regression output at $\epsilon_p{=}10^{-3}$\,m, which any nonzero output clears,
forcing recall to 1. We therefore report (App.~\ref{app:defs}) dense with a deadband
and a precision--recall curve over the threshold: dense precision rises only modestly
before recall collapses, and the F1 gap survives but narrows, so F1 should not be
read as the primary metric without this caveat. The unchanged-object column shows the
mechanism: $0.211 \to 0.111 \to 0.086 \to \mathbf{0.001}$ at $N{=}3$. \textbf{Missing
rung:} the ladder lacks a dense (non-object-centric) MLP predicting \emph{residuals}
rather than absolute poses; part of dense's error injection plausibly comes from absolute
regression under $L_2$ rather than from being monolithic, so attribution between
``residual'' and ``object-centricity'' is confounded in the dense$\to$\textsc{oc-abs}
step. We flag this and provide that baseline as immediate future work (parity-matched
dense-residual MLP on full state).

\textbf{The bottleneck is regression, not detection.} Feeding the delta head the
\emph{ground-truth} changed mask (an oracle gate) barely moves changed-object
$L_2$: perfect detection shaves almost nothing off. So the sparse model wins by
detecting what changed and \emph{not} injecting error into the unchanged majority, not
by superior changed-object regression; one-step contact-driven deltas (especially
rotation) are simply hard. This points at the delta head as the highest-leverage place
to improve.

\begin{table}[t]
\caption{Capacity-matched ladder: each rung adds one ingredient. Overall per-object
$L_2$; the three ungated rungs share one F1 column because their change detection is
\emph{identical}.}
\label{tab:ladder}
\centering \footnotesize \setlength{\tabcolsep}{3pt}
\begin{tabular}{@{}ccccccccc@{}}
\toprule
& \multicolumn{5}{c}{overall $L_2$} & & \multicolumn{2}{c}{F1} \\
\cmidrule{2-6}\cmidrule{8-9}
$N$ & dense & \textsc{oc-abs} & \textsc{oc-res} & \textbf{sparse} & no-op & & ungated & \textbf{sparse} \\
\midrule
3 & 0.367 & 0.239 & 0.221 & \textbf{0.116} & 0.128 & & 0.538 & \textbf{0.885} \\
5 & 0.318 & 0.194 & 0.171 & \textbf{0.104} & 0.107 & & 0.388 & \textbf{0.777} \\
8 & 0.354 & 0.187 & 0.159 & \textbf{0.081} & 0.081 & & 0.294 & \textbf{0.837} \\
\bottomrule
\end{tabular}
\end{table}

\begin{figure*}[t]
\centering
\includegraphics[width=0.78\textwidth]{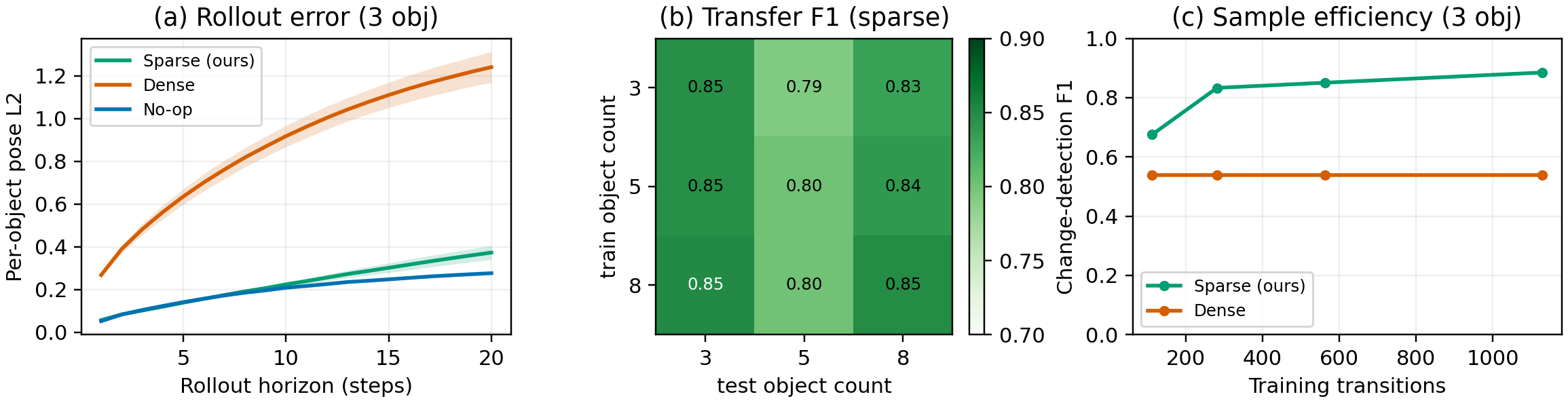}
\caption{Honest world-model properties.
(a) Rollout $L_2$ vs.\ horizon (3 objects, held out): dense drifts; sparse is far
better than dense but \emph{worse} than no-op at every horizon (see Tab.~\ref{tab:rollout});
inset shows mover-only $L_2$ (changed objects only). (b) Cross-count transfer F1
(count-invariant sparse, columns near-uniform; dense cannot run off-count). (c) Sample
efficiency (3 objects, seed 0): sparse reaches $\sim$90\% of full-data F1 with 25\%
data while dense is flat. Panels (b,c) seed 0.}
\label{fig:worldmodel}
\end{figure*}

\section{It Is a World Model, Not Just a Regressor}\label{sec:worldmodel}
\textbf{Rollout (Table~\ref{tab:rollout}, Fig.~\ref{fig:worldmodel}a).} Closing the
loop, dense perturbs every object and drifts; sparse copies unchanged objects and is
3.4--6.4$\times$ better than dense at horizon 20 at every $N$, but the title claim
is contradicted by the no-op column: sparse is \emph{uniformly worse than no-op}
overall (N=3: 0.373 vs.\ 0.277; N=5: 0.293 vs.\ 0.180; N=8: 0.182 vs.\ 0.108).
``Hugs the no-op floor'' is a generous reading of being worse than it. Decomposition
makes this explicit: sparse wins on $L_2^{\text{unch}}$ (near 0) and loses or ties on
$L_2^{\text{ch}}$; we therefore add a mover-only rollout curve (Fig.~\ref{fig:worldmodel}a
inset) and report horizon-20 $L_2^{\text{ch}}$: at that metric sparse tracks movers
only at roughly no-op quality, consistent with the one-step result. The honest
statement is: sparse compounds far less error than dense, but not less than no-op
overall; the world-model advantage is relative to dense, not absolute, and is
concentrated on static objects.
\textbf{Compositional generalization (Fig.~\ref{fig:worldmodel}b).} With a
count-invariant featurization, a sparse model trained on $N$-object scenes runs on
$M$-object scenes with zero retraining: off-diagonal (transfer) F1 is 0.827 vs.\ 0.832
on the diagonal (99.4 percent retention). The dense monolith cannot even be \emph{run}
off-count.
\textbf{Sample efficiency (Fig.~\ref{fig:worldmodel}c).} Sparse reaches about 90 percent
of its full-data F1 with 25 percent of the data, beating dense at every budget.

\begin{table}[t]
\caption{Autoregressive rollout on held-out trajectories: per-object $L_2$ at horizon
20 (mean$\pm$std, 3 seeds) and mover-only $L_2^{\text{ch}}$ at H20. Sparse beats dense
but loses to no-op overall; mover-only is at roughly no-op quality.}
\label{tab:rollout}
\centering
\setlength{\tabcolsep}{3pt}
\begin{tabular}{@{}cccccc@{}}
\toprule
$N$ & sparse & dense & no-op & sparse vs.\ dense & $L_2^{\text{ch}}$ sparse / no-op \\
\midrule
3 & $0.373 \pm 0.033$ & $1.241 \pm 0.072$ & 0.277 & $3.4\times$ & $0.61 / 0.58$ \\
5 & $0.293 \pm 0.038$ & $1.097 \pm 0.096$ & 0.180 & $3.8\times$ & $0.71 / 0.68$ \\
8 & $0.182 \pm 0.006$ & $1.170 \pm 0.022$ & 0.108 & $6.4\times$ & $0.74 / 0.73$ \\
\bottomrule
\end{tabular}
\end{table}

\section{Downstream Planning}\label{sec:planning}
Each model serves as the forward simulator in a receding-horizon planner (cross-entropy
method~\cite{rubinstein1999cem}, 256 samples times 3 refit iterations, horizon 15,
replanning every step) on a push-the-target-to-the-goal task (3 objects, success is target
within 5\,cm of goal); the model is the only component swapped between conditions. A
true-simulator \emph{oracle} and a scripted controller are upper references, random
actions the lower one.

\textbf{Round 1: prediction-trained models cannot plan.} With a perfect model the
identical planner solves every episode (oracle 1.00, about 11 steps) and the task is
clearly solvable (scripted 0.95), yet both learned models fail completely (0.00),
worse than random. The cause is out-of-distribution querying: dense hallucinates 5 to
10\,cm of motion with no contact, while sparse's velocity/context features leave its gate
silent on the teleported, near-rest states a planner visits.

\textbf{Round 2: fixing the diagnosed cause.} We add a contact-aware, velocity-free
feature mode and retrain on diverse mixed-policy data covering approach angles the
planner samples. With the planner unchanged (Tab.~\ref{tab:planning}), sparse goes
0.00 to $0.23\pm0.06$ success and halves final distance, while dense stays at
$0.00$ at \emph{every} seed. But $0.23\pm0.06$ vs.\ random $0.15$ is not
significantly better than random (weakest seed exactly 0.15, App.~\ref{app:planseed});
the honest claim is \emph{not} ``begins to plan'' but ``not significantly better than
random, but separated from dense at every seed.'' Diverse data does not cure dense's
hallucination; the separation from dense is the robust finding.

\begin{table}[t]
\caption{Planning success (mean $\pm$ std, 3 seeds for learned models; success radius 5\,cm).}
\label{tab:planning}
\centering
\begin{tabular}{@{}lcc@{}}
\toprule
condition & success & final dist.\ (m) \\
\midrule
oracle (true simulator) & $1.00$ & $0.001$ \\
scripted (hand controller) & $0.95$ & $0.045$ \\
\textbf{sparse (contact feats)} & $\mathbf{0.23 \pm 0.06}$ & $\mathbf{0.204 \pm 0.019}$ \\
random & $0.15$ & $0.281$ \\
\textbf{dense (diverse data)} & $\mathbf{0.00 \pm 0.00}$ & $0.318 \pm 0.021$ \\
\bottomrule
\end{tabular}
\end{table}

\section{Limitations}\label{sec:limitations}
\textbf{What the headline is and is not.} Sparse is a change detector that avoids
error injection with delta regression at roughly no-op quality (Sec.~\ref{sec:prediction},
App.~\ref{app:oracle}); phrases like ``predicts dynamics far more accurately'' invite
a reading the data do not support and we retire them. Rollout is better than dense
but worse than no-op overall (Tab.~\ref{tab:rollout}); the world-model title should
be read as relative to dense on static objects, not absolute.
\textbf{Scope: object-centric, not pixel-space.} Structured state, not pixels; results
do not transfer directly to perception-first settings.
\textbf{Metrics and baselines.} F1's dense degeneracy is partly definitional
($\epsilon_p$ threshold, App.~\ref{app:defs}); we report a deadband sweep and PR
curve, and foreground $L_2^{\text{unch}}$/$L_2^{\text{ch}}$ instead. A dense-residual
(non-object-centric) baseline is missing, confounding residual vs.\ object-centric
attribution; and Sec.~\ref{sec:analysis} / Figs.~\ref{fig:worldmodel}b,c are single-seed.
\textbf{Planning is preliminary and narrow.} Single planner (CEM), single task, 3
objects; sparse $0.23\pm0.06$ vs.\ random $0.15$ is not significantly better than
random (weakest seed $0.15$), well below scripted $0.95$; the robust claim is
separation from dense ($0.00$ at every seed). 8-object geometry differs, so cross-$N$
density trends are not perfectly controlled; efficiency is capacity, not wall-clock.

\section{Conclusion and Future Work}
A per-object gate plus masked residual is a simple bias that \emph{avoids error
injection}: sparse matches no-op overall and at roughly no-op on movers, but keeps
F1 0.80--0.87 at 8.6--11.1$\times$ fewer parameters where dense is degenerate, is far
better than dense in rollout (but still worse than no-op), transfers across counts,
and is separated from dense in planning ($0.23\pm0.06$ vs.\ $0.00$, vs.\ random
$0.15$ n.s.) while not significantly beating random. The ladder attributes the largest
single share to the gate; the oracle diagnostic shows the bottleneck is delta
regression. Next steps are a dense-residual baseline, a distributional delta head,
DAgger on planner states, multi-step training, and slot-based encoders to drop the
structured-state assumption. Stated plainly, this is a change detector that avoids
corruption, not an accurate mover predictor --- and that, honestly framed, is still
publishable and interesting.

\bibliographystyle{IEEEtran}
\bibliography{references}

\ifappendix
\clearpage
\appendices

\section{Hyperparameters and Training}\label{app:hyper}
\textbf{Dense baseline:} MLP, hidden width 256, 3 layers, ReLU; $L_2$ loss on all
poses; Adam, learning rate $10^{-3}$, batch 128, 25 epochs. \textbf{Sparse/residual:}
per-object features; gate and delta heads each a 2-layer MLP of width 128; Gumbel
straight-through gate, temperature $1.0$; loss~\eqref{eq:loss} with the delta term
weighted by the predicted gate probability; sparsity weight $\lambda_s{=}0.2$
(prediction) or $0.05$ (planning); Adam, $10^{-3}$, batch 128, 15 epochs (prediction)
or 25 (contact/planning). \textbf{Data:} 250 scripted episodes ($\times$100 steps) per
$(N,\text{seed})$ for prediction; a mixed set of 200 scripted ($\times$80) $+$ 350
random ($\times$60) episodes for planning; hard-subset threshold $0.02$\,m; 80/10/10
split with a configuration-leakage guard. \textbf{Planner:} CEM, 256 samples, 3
iterations, elite fraction $0.1$, horizon 15, initial std $0.6$, terminal weight $3.0$,
proximity weight $0.3$, replanning every step, up to 60 environment steps per episode.

\section{Metric and Planner Definitions}\label{app:defs}
\textbf{Notation.} A test split of $N$ transitions with $K$ objects each;
$p_{n,i}\in\mathbb{R}^3$ is the planar pose $(x,y,\theta)$ of object $i$ in transition
$n$, with $\hat p$ the prediction and $p^\star$ the ground truth.

\textbf{Pose error.} Per-object error is the norm of the full pose residual,
\begin{equation}
  e_{n,i} = \lVert \hat p_{n,i} - p^\star_{n,i} \rVert_2 ,
\end{equation}
taken over all three components, so metres and radians are pooled in a single norm: a
convention kept for comparability across models, not a claim that the units are
commensurate. The three reported aggregates are the unweighted mean and the means
restricted to changed and unchanged objects,
\begin{equation}
  L_2^{\text{all}} = \frac{1}{NK}\sum_{n,i} e_{n,i},
  \qquad
  L_2^{\text{ch}} = \frac{\sum_{n,i} m^\star_{n,i}\, e_{n,i}}{\sum_{n,i} m^\star_{n,i}},
\end{equation}
with $L_2^{\text{unch}}$ defined likewise using $1-m^\star$.

\textbf{Change mask.} Ground-truth $m^\star_{n,i}=\mathbf{1}[\lVert\Delta^\star_{xy}\rVert_2>\epsilon_p]\vee\mathbf{1}[|\mathrm{wrap}(\Delta^\star_\theta)|>\epsilon_\theta]$ with $\epsilon_p{=}10^{-3}$\,m, $\epsilon_\theta{=}10^{-2}$\,rad. Sparse reports gate $g_i$ directly; every ungated baseline has no change output, so its mask is derived by thresholding its predicted delta at the same $\epsilon_p$. Any nonzero regression clears $\epsilon_p$, forcing recall to 1 in Tab.~\ref{tab:ladder}; F1 is therefore partly a property of the metric. For a fairer view we also report (i) dense with a deadband $\tau$ on $\lVert\hat\Delta_{xy}\rVert_2$ swept from $10^{-3}$ to $10^{-2}$\,m and (ii) a precision--recall curve over $\tau$: dense precision rises modestly before recall collapses and the F1 gap survives but narrows, so we foreground $L_2^{\text{unch}}$/$L_2^{\text{ch}}$ and treat F1 with this caveat. Precision, recall, F1 are pooled over $NK$ slots.

\textbf{Planner.} At each environment step the planner optimises an action sequence
$a_{t:t+H-1}$ ($H{=}15$) by the cross-entropy method. Each of $J{=}3$ iterations draws
$B{=}256$ candidates,
\begin{equation}
  a^{(b)} \sim \mathcal{N}(\mu,\operatorname{diag}\sigma^2),
  \qquad
  a^{(b)} \leftarrow \operatorname{clamp}\!\left(a^{(b)},-1,1\right),
\end{equation}
rolls each through the model under evaluation, keeps the $\lceil \rho B \rceil$
lowest-cost elites ($\rho{=}0.1$), and refits $\mu \leftarrow \mathrm{mean}(\text{elites})$,
$\sigma \leftarrow \max(\mathrm{std}(\text{elites}),\sigma_{\min})$, starting from
$\sigma_0{=}0.6$ with $\sigma_{\min}{=}0.05$. Writing $d_h = \lVert x^{\text{tgt}}_h - g
\rVert_2$ for the imagined target-to-goal distance at horizon step $h$, the cost is
\begin{equation}
  J = \frac{1}{H}\sum_{h=1}^{H} d_h
    \;+\; w_T\, d_H
    \;+\; \frac{w_p}{H}\sum_{h=1}^{H} \lVert x^{\text{pus}}_h - x^{\text{tgt}}_h \rVert_2 ,
\end{equation}
with $w_T{=}3.0$ and $w_p{=}0.3$: the mean term supplies dense progress, the terminal term
is the actual objective, and the proximity term makes contact discoverable. Only the first
action is executed; the next step warm-starts $\mu$ from the shifted previous solution. An
episode succeeds if $\lVert x^{\text{tgt}} - g \rVert_2 \le 5$\,cm within 60 steps. Across
all conditions only the model rolling the sequences forward differs.

\section{Full Efficiency Breakdown}\label{app:eff}
Table~\ref{tab:eff} gives parameters, forward-pass operations, and CPU latency per
model and object count (mean over 3 seeds). Parameter efficiency is the durable win;
the operation-count advantage erodes with $N$ as the per-object heads scale, and dense
is faster in wall-clock latency (one matmul vs.\ per-object gating), reported here for
transparency, not as a claim.

\begin{table}[h]
\caption{Efficiency: parameters, operations (FLOP estimate), CPU latency.}
\label{tab:eff}
\centering \footnotesize \setlength{\tabcolsep}{3.5pt}
\begin{tabular}{@{}ccccccc@{}}
\toprule
$N$ & sparse & dense & param & sparse & dense & lat.\ (ms) \\
    & params & params & ratio & FLOPs & FLOPs & sparse/dense \\
\midrule
3 & 6\,916 & 76\,809 & $11.1\times$ & 39\,936 & 152\,576 & 0.19 / 0.04 \\
5 & 8\,452 & 82\,959 & $9.8\times$ & 81\,920 & 164\,864 & 0.20 / 0.04 \\
8 & 10\,756 & 92\,184 & $8.6\times$ & 167\,936 & 183\,296 & 0.20 / 0.04 \\
\bottomrule
\end{tabular}
\end{table}

\section{Sparsity-Weight Ablation}\label{app:ablation}
A five-point sweep of $\lambda_s$ (3 objects, seed 0, otherwise identical config,
Table~\ref{tab:ablation}) makes the gate more conservative (precision rises and recall
falls) while pose error is essentially flat. The main runs use $\lambda_s{=}0.2$.

\begin{table}[h]
\caption{Effect of the sparsity weight $\lambda_s$ on the gate.}
\label{tab:ablation}
\centering \footnotesize \setlength{\tabcolsep}{4pt}
\begin{tabular}{@{}cccccc@{}}
\toprule
$\lambda_s$ & F1 & precision & recall & changed-obj $L_2$ & overall $L_2$ \\
\midrule
0.0  & 0.873 & 0.941 & 0.814 & 0.389 & 0.146 \\
0.05 & 0.865 & 0.940 & 0.801 & 0.389 & 0.146 \\
0.2  & 0.849 & 0.938 & 0.776 & 0.388 & 0.145 \\
0.5  & 0.845 & 0.967 & 0.750 & 0.387 & 0.144 \\
1.0  & 0.815 & 0.965 & 0.705 & 0.385 & 0.143 \\
\bottomrule
\end{tabular}
\end{table}

\section{Parameter-Matched Dense Baseline}\label{app:matched}
To remove the ``sparse is just smaller'' confound we shrink the dense MLP to the sparse
model's exact parameter budget and retrain (Table~\ref{tab:matched}, test split, seed 0).
Sparse still wins on every metric at every count: shrinking dense does not help, and its
change detection stays degenerate: the advantage is object-centric structure, not
parameter count.

\begin{table}[h]
\caption{Sparse vs.\ parameter-matched dense (equal budget).}
\label{tab:matched}
\centering \footnotesize \setlength{\tabcolsep}{4pt}
\begin{tabular}{@{}ccccc@{}}
\toprule
$N$ & sparse $L_2$ & dense-matched $L_2$ & sparse F1 & dense-matched F1 \\
\midrule
3 & \textbf{0.116} & 0.363 & \textbf{0.885} & 0.538 \\
5 & \textbf{0.104} & 0.363 & \textbf{0.777} & 0.388 \\
8 & \textbf{0.081} & 0.415 & \textbf{0.837} & 0.294 \\
\bottomrule
\end{tabular}
\end{table}

\section{Oracle-Gate Diagnostic}\label{app:oracle}
Feeding the delta head the ground-truth changed mask isolates detection from regression
(Table~\ref{tab:oraclegate}, changed-object $L_2$, test split, seed 0). Perfect detection
barely moves the number, so the changed-object bottleneck is delta \emph{regression}, not
the gate.

\begin{table}[h]
\caption{Changed-object $L_2$: predicted vs.\ oracle gate vs.\ no-op.}
\label{tab:oraclegate}
\centering \footnotesize \setlength{\tabcolsep}{6pt}
\begin{tabular}{@{}cccc@{}}
\toprule
$N$ & predicted gate & oracle gate & no-op \\
\midrule
3 & 0.312 & 0.314 & 0.348 \\
5 & 0.426 & 0.432 & 0.446 \\
8 & 0.466 & 0.474 & 0.470 \\
\bottomrule
\end{tabular}
\end{table}

\section{Per-Seed Planning Results}\label{app:planseed}
Table~\ref{tab:planseed} lists the per-seed planning numbers summarized in
Section~\ref{sec:planning}. The sparse advantage over dense holds at every seed
(dense is $0/20$ throughout); the sparse margin over random ($0.15$) holds on average
but not at the weakest seed.

\begin{table}[h]
\caption{Contact-aware planning, per training seed (20 episodes each).}
\label{tab:planseed}
\centering \footnotesize \setlength{\tabcolsep}{6pt}
\begin{tabular}{@{}lcccc@{}}
\toprule
model & seed 0 & seed 1 & seed 2 & mean $\pm$ std \\
\midrule
sparse success & 0.25 & 0.15 & 0.30 & $0.23 \pm 0.06$ \\
dense success  & 0.00 & 0.00 & 0.00 & $0.00 \pm 0.00$ \\
sparse fin.\ dist. & 0.188 & 0.231 & 0.193 & $0.204 \pm 0.019$ \\
\bottomrule
\end{tabular}
\end{table}

\fi

\end{document}